\documentclass[10pt,twocolumn,letterpaper]{article}

\pdfoutput=1
\usepackage{cvpr}
\usepackage{times}
\usepackage{epsfig}
\usepackage{graphicx}
\usepackage{amsmath}
\usepackage{amssymb}

 \hypersetup{
   breaklinks=true,
   bookmarks=false,
   colorlinks   = true, %Colours links instead of ugly boxes
   urlcolor     = blue, %Colour for external hyperlinks
   linkcolor    = red, %Colour of internal links
   citecolor   = green %Colour of citations
}

\usepackage{xcolor, soul}
\sethlcolor{green}

\cvprfinalcopy % *** Uncomment this line for the final submission

\def\cvprPaperID{1} % *** Enter the CVPR Paper ID here

\begin{document}

%%%%%%%%% TITLE
\title{Utilizing Mask R-CNN for Waterline Detection in Canoe Sprint Video Analysis}

\author{Marie-Sophie von Braun\\ Laboratory for Biosignal Processing\\ Leipzig University of Applied Sciences\\ {\tt\small msvbraun@gmail.com}
\and Patrick Frenzel\\ Laboratory for Biosignal Processing\\ Leipzig University of Applied Sciences\\ {\tt\small patrick.frenzel@htwk-leipzig.de}
\and Christian Käding\\ Institute for Applied Training Science Leipzig\\ {\tt\small kaeding@iat.uni-leipzig.de}
\and Mirco Fuchs\\ Laboratory for Biosignal Processing\\ Leipzig University of Applied Sciences\\ {\tt\small mirco.fuchs@htwk-leipzig.de}
}

\maketitle
%\thispagestyle{empty}

%%%%%%%%% ABSTRACT
\begin{abstract}
  Determining a waterline in images recorded in canoe sprint training is an important component for the kinematic parameter analysis to assess an athlete's performance. Here, we propose an approach for the automated waterline detection. First, we utilized a pre-trained Mask R-CNN by means of transfer learning for canoe segmentation. Second, we developed a multi-stage approach to estimate a waterline from the outline of the segments. It consists of two linear regression stages and the systematic selection of canoe parts. We then introduced a parameterization of the waterline as a basis for further evaluations. Next, we conducted a study among several experts to estimate the ground truth waterlines. This not only included an average waterline drawn from the individual experts annotations but, more importantly, a measure for the uncertainty between individual results. Finally, we assessed our method with respect to the question whether the predicted waterlines are in accordance with the experts annotations. Our method demonstrated a high performance and provides opportunities for new applications in the field of automated video analysis in canoe sprint.
  %The waterline approximates a boundary between the blurred water surface and the body of the canoe.
\end{abstract}

%%%%%%%%% BODY TEXT
\section{Introduction}
Recording and analysing video sequences is a common method for the quantification, logging and optimization of the technique of athletes performing canoe and kayak sprint~\cite{McDonnell2012, Michael2009}. A particularly important form is the recording from the position of a motorboat that moves in parallel direction to the canoe~\cite{Robinson2011, Tay2018}. While moving at the same speed, the athlete is recorded from an approximately perpendicular perspective with respect to the movement direction. This ensures standardized recording conditions which are the basis to assess the performance of athletes and in particular of their technique. The actual analysis is then based on determining kinematic parameters and their comparison to well known target values.

\begin{figure}[t]
\begin{center}
\includegraphics[width=\linewidth]{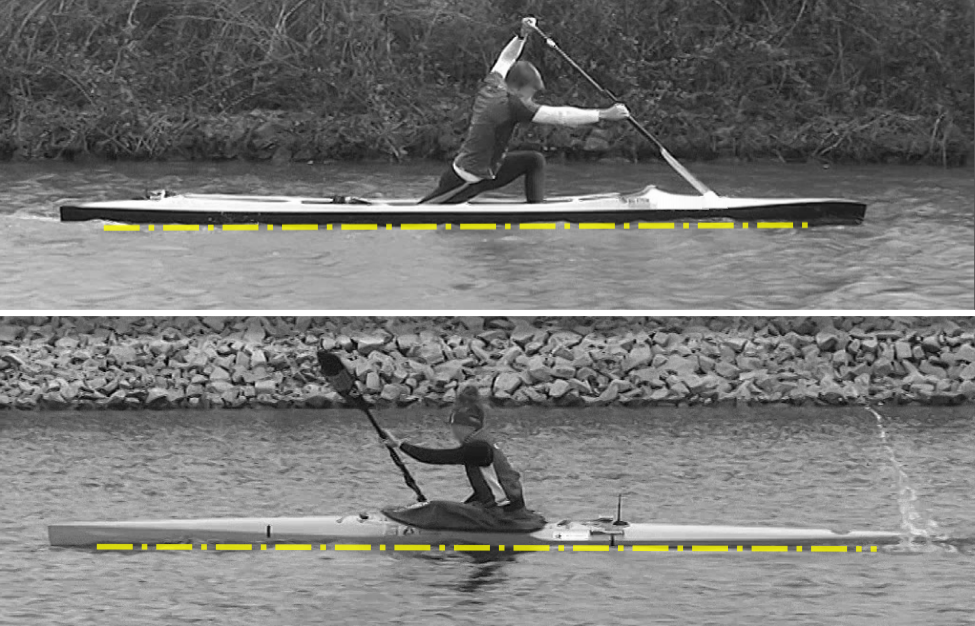}
\end{center}
\caption{Illustration of the disciplines canoe sprint (top) and kayak sprint (bottom). The prediction of the waterline (dash-dotted yellow line) is crucial for the analysis of kinematic parameters to assess the technique of athletes.}
\label{fig:disciplines}
\end{figure}

The underlying recording conditions in the free environment are subject to variations (\eg due to water movement, hand-held camera, inherent cyclic speed changes during paddling) that affect the comparability of the analyses performed on these recordings. Hence, an analysis is usually restricted to a narrow time range to minimize potential variations, \ie often only a few or even a single paddle cycle. The technique is then analyzed based on several single images which are selected in accordance with a paddle phase model that defines the beginning, end and intermediate stage of a cycle~\cite{Mann1983, Herrmann1979, Michael2009}. The parameters determined in each of these images allow a comparison between multiple athletes as well as between repeated training runs of the same athlete.

The analysis of these single images aims at measuring several distances and angles, \ie the kinematic parameters, in the projected 2D image plane~\cite{Englert2007}. It comprises the identification of several landmarks as, in particular, key-points on the body of the athlete, a so called paddle line, that means a straight line through the longitudinal middle of the paddle shaft, and a waterline. As shown in Fig.~\ref{fig:disciplines}, the latter approximates a boundary between the blurred water surface and the more well-defined hull of the canoe. The waterline is of particular interest for two reasons. First, it serves as a reference with respect to the kinematic parameters to be estimated. Second, its exact position and orientation is often difficult to specify in practice. Turbulences in water, waves and splashes, reflections, camera jitter, a varying camera position or even poor image quality give rise to ambiguity in this task.

The manual analysis procedure of these standardized video sequences is time consuming, requires expert knowledge and is also subject to individual errors. There is a general interest in an automatic procedure which provides a much faster analysis and which is also less prone to individual errors. The advances in image processing by means of deep neural networks in recent years pave the way for the development of a widely automated analysis of video recordings in canoe sprint. On the one hand, the rise of human 2D pose estimation algorithms~\cite{cao2016openpose, Cheng2019, Xiao2018, Kocabas2018} and their task-specific optimization provides the opportunity to automatically determine key-points on an athlete's body in a given image~\cite{Huang2019}. The application of such algorithms for key-point detection has proven effective in a variety of applications such as, \eg, skeleton tracking of players in sports~\cite{Bridgeman2019}, swimming style classification~\cite{Einfalt2018} and stroke frequency detection~\cite{Victor2017}, and pose mining in long jump~\cite{Lienhart2018}. There is reason to expect them to work in canoe sprint analysis as well. In fact, their potential use for video analysis in canoe sprint has recently been reported~\cite{Fuchs2018}. On the other hand, approaches for the pixel-wise segmentation of objects in an image such as Mask R-CNN~\cite{He2017MaskR} might serve as a basis for an automated detection of the canoe and the paddle which can subsequently be used to determine the waterline and the paddle line, respectively. However, the prospects of success are much less obvious compared to pose estimation.

This work presents an approach for an automated detection of the waterline. Our method is based on image segmentation using the Mask R-CNN network pre-trained on the COCO dataset~\cite{Lin2014} and a subsequent multi-stage procedure that includes two linear regression steps. As to the mentioned uncertainty of waterlines defined by several experts, we conducted an evaluation study and derived a gold standard to assess the predicted waterlines.

The contributions of this paper include (1) an adoption of Mask R-CNN for canoe segmentation and discrimination of the disciplines canoe sprint and kayak sprint (compare Fig.~\ref{fig:disciplines}), (2) a procedure to estimate a waterline given the segmented shape of the canoe, (3) a gold standard to assess predicted waterlines with respect to human experts and (4) a performance analysis of the proposed waterline detection method.

%------------------------------------------------------------------------
\section{Related Work}
\subsection{Mask R-CNN in Sports Applications}
Since its superiority in the instance segmentation task of the COCO Challenge 2017, Mask R-CNN has been widely used for scene analyzing in sports videos. The applications range from ball detection~\cite{buric2018ball} to jersey number recognition~\cite{Liu_2019_CVPR_Workshops}, and further to player tracking~\cite{pobar2019player} and events identification~\cite{Kazi2019events, pobar2018actions}.

Challenges in these applications are amongst others the dynamics of the subject and the numerous occlusions of the tracked object that occur during the game. When analyzing canoe sprint videos though, there is an additional challenge: the peculiarity of the medium water, which render a robust and accurate detection of the waterline difficult.

\subsection{Waterline Problem}
Several works focus on estimating the waterline on images or videos of rivers or lakes in order to detect the sailing area of an autonomous boat. Wei and Zhang~\cite{wei2016waterline} present a waterline detection method based on texture analysis of river images with local binary patterns (LBPs) and gray level co-occurrence matrix (GLCM). Steccanella \etal~\cite{steccanella2018waterline} apply a supervised approach based on a Fully Convolutional Neural Network for obtaining a pixel-wise image segmentation.

These methods however rely on a detection of the water area and focus on its boundary with the horizon line. Hence, they cannot be used to estimate the separation line between the lower part of a canoe hull and the water surface.
Our case requires a segmentation of the canoe hull within the water body. The use of Mask R-CNN for this case is the subject of several papers~\cite{schweitzer2018aerial, zhang2018vessel, nie2018ship, you2019ship, nie2019ship, nie2020ship}. However, these approaches are applied exclusively to satellite images. Due to the aerial perspective and their remote sensing character, these are not comparable to the video sequences of canoe sprints that are examined here.

%------------------------------------------------------------------------
\section{Proposed Approach}
\subsection{Task definition}
The goal of our work is to determine a waterline in an RGB image $I\in \mathbb{R}^{m\times n \times 3}$ drawn from video sequences (50 frames per second) in the disciplines canoe sprint and kayak sprint (see Fig.~\ref{fig:disciplines}). Here, the spatial resolution is $m=1024~\times~n=576$ pixels. It is assumed that the images are recorded from an approximately perpendicular perspective with respect to the movement direction of the canoe. Moreover, only minor variations of the distance and the relative position between canoe and motorboat are expected for consecutive images selected from the short time window to be analyzed in the training run of an athlete.

The determination of the waterline is a regression problem. The goal is to approximate a straight line that separates the visible part of the canoe from the invisible part below the water surface. Waves, splashes and other disturbance that might occlude the canoe hull must be considered when approximating the line. The linear approximation of the waterline should be particularly accurate in the central part of the canoe right below the athlete, since this segment is notably important for the subsequent derivation of the kinematic parameters.

We propose a two-staged approach for waterline prediction. First, it is based on a pixel-wise segmentation of the canoe by means of a Mask R-CNN that we adjusted to this particular task using transfer learning. This is presented in Sec.~\ref{ss:rcnn}. Second, it employs a multi-stage procedure to confine the pixels of the canoe segmentation to those close to the water surface which can finally be used to define a waterline. This is shown in Sec.~\ref{ss:waterline}.

\subsection{Canoe Segmentation with Mask R-CNN}
\label{ss:rcnn}
\subsubsection{Method}
The first stage of our approach is a pixel-wise instance segmentation of each canoe object contained in the image. The Mask R-CNN method proposed by He \etal~\cite{He2017MaskR} has evolved to a state of the art approach for pixel-wise instance segmentation. It is a two-stage framework built on top of a Faster R-CNN~\cite{ren2015fasterrcnn}: the first stage generates object proposals, while the second stage predicts the class of each object, refines its bounding box and generates a corresponding binary mask on pixel level. Both stages are connected to a backbone, in our case a ResNet101~\cite{He2016} paired with a Feature Pyramid Network, that serves as a feature extractor. Hence, such a net is able to detect the set of objects $\Omega$, with its elements $\omega_v=\left(M_v, c_v, p_v \right) \in \Omega, v\in\mathbb{N}$ defining the class $c_v$, its confidence value $p_v \in \mathbb{R}, 0 < p_v \leq 1$ and its binary mask $M_v \in \{0,1\}^{m \times n}$ of an instance. The latter provides a pixel-wise binary appearance of an object in an image.

The particular segmentation problem in canoe sprint video analysis is highly specific. That means that the algorithm is not required to identify any additional object in an image but only the canoe of an athlete. We exploited this fact and restricted the potential output objects to canoes used in canoe  sprint and kayak sprint (\ie, $c_v \in [{\mathit{canoe},\mathit{kayak}}]$), both of which are actually canoes but with a slightly different appearance. The segmented canoe as defined by its binary segmentation mask $M_v$ can subsequently be used to determine the lower part of the outline of the canoe.

\subsubsection{Dataset and Implementation Details}
We adopted the Mask R-CNN implementation as provided by Matterport~\cite{matterport_maskrcnn_2017} which employs a ResNet101 architecture as a backbone~\cite{He2016}. It is implemented in TensorFlow~\cite{abadi2016tensorflow} and Keras~\cite{chollet2015keras}. As described above, we restricted the output layer for the segmentation to only two types of objects, \ie canoes for the disciplines canoe sprint and kayak sprint. We applied transfer learning to train our model. Therefore, we used the pre-trained weights that resulted from training the Matterport implementation on the COCO dataset~\cite{Lin2014}. We used the following training parameters: 400 iterations, 300 steps per iteration, SGD solver, learning rate $0.001$, momentum $0.9$, one image batch size and weight decay $0.0001$.

We carried out image annotation to derive training and test sets as follows. Given were a total number of 66 video sequences from both disciplines from which 250 images were randomly selected. We used the VGG Image Annotator~\cite{dutta2019vgg, dutta2016via} to define polygones that mimic the canoe hull in each image. Next, we used 210 images (58 from canoe sprint, 152 from kayak sprint) to define a training set and 40 for the validation set (11 canoe sprint, 29 kayak sprint). Moreover, we selected 30 of these images for the validation set in a way that ensured that they belong to video sequences from which no other image is used for training. In case of the other 10 images, the canoes they contain are already known to the model due to the training process.

During training, the image was either kept in its original form ($p=0.5$) or processed using data augmentation as follows: flipping in horizontal direction ($p=0.5$), rotation by 2 degree and cropping/padding in a range from -15\,\% to 15\,\% in both image dimensions. The former represents a zoom-in effect, the latter corresponds to zooming out.
 %augmentation = iaa.Sometimes(0.5, [
    %iaa.Fliplr(0.5),
    %iaa.Affine(rotate=2),
    %iaa.CropAndPad(percent=(-0.15, 0.15), pad_mode="constant")
    %])

\subsection{Waterline Detection}
\label{ss:waterline}
\begin{figure}[t]
\begin{center}
\includegraphics[width=\linewidth]{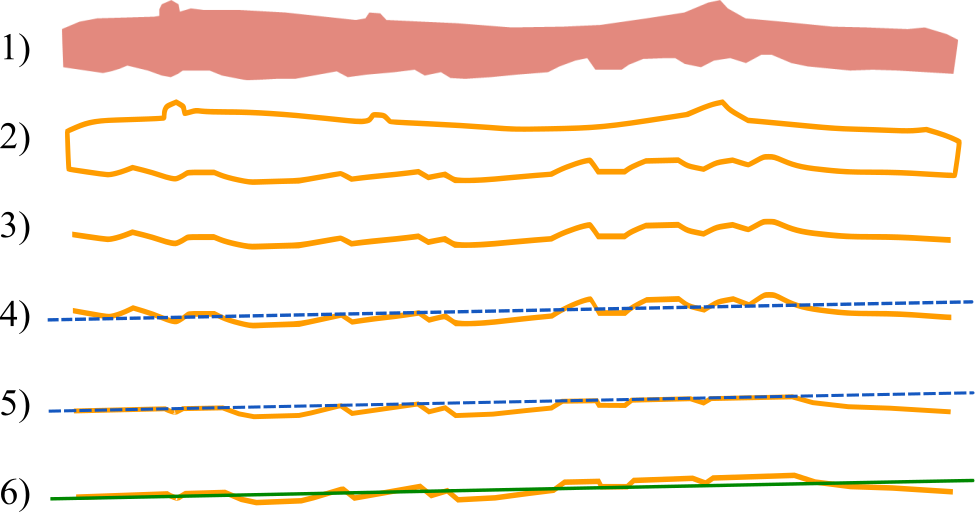}
\end{center}
   \caption{Illustration of the iterative procedure to predict a waterline on the basis of a canoe segmentation. Details for steps 1-6 are provided in the text. The dashed blue lines in steps 4 and 5 are the same. The solid green line in step 6 represents the predicted waterline. It corresponds to the waterlines illustrated in Fig.~\ref{fig:disciplines}.}
\label{fig:waterline_alg}
\end{figure}

The second stage of our approach comprises an iterative procedure to derive a waterline given the binary segmentation mask $M \in \{0,1\}^{m \times n} $ as provided by the segmentation approach presented in the previous section. The procedure is depicted in Fig.~\ref{fig:waterline_alg}, step 1 shows the initial segmentation. First, all points $C \subset	M$ that represent the contour of the canoe are determined (step 2). As a result, $C$ contains at least two tuples for each image coordinate $x_i$ where the canoe segment was found, \ie ($x_i, y_1$) and ($x_i, y_j$), $j\in\mathbb{N}, j\geq 2$, all of which belong to the contour. It is obvious that the waterline is close to the bottom of the canoe and we therefore reject the tuples belonging to the upper part and keep the others. Since we defined $y=0$ at the top of the image, the tuples with the larger coordinate, \ie $\mathit{max}\left(y_1,...,y_j\right)$, are used for further processing, leading to $C'$. Hence, $C' \subset C \subset M$ defines the set of points belonging to the bottom line of the canoe contour (step 3). Third, a linear regression $\mathcal{L}_1\left(C'\right)$ is performed on the set of points contained in $C'$ (step 4). All points above this regression line are subsequently removed to form $C'' \subset C'$ (step 5). This step mimics cropping of small waves and splashes. Finally, another linear regression step $\mathcal{L}_2\left(C''\right)$ is performed on the set of points in $C''$ (step 6). Its result defines the predicted waterline.
%
%- from segmentation to waterline
%* Mask R-CNN detects canoe mask (usually only 1 canoe per image)
%* OpenCV extracts outline from mask (find\_contours from skimage) --> smootness filter??
%* extract subset of outline’s bottom line
%* first simple linear regression on these points
%* keep only subset of points under or on the regression line
%* second simple linear regression on these points
%* 3 pixel shift downwards based on empirical observations
%-> final detected waterline

%------------------------------------------------------------------------
\section{Evaluation}
The evaluation of predicted waterlines requires reference data to assess its accuracy. However, quantitative reference data, for example, derived from passive optical markers does not exist. Moreover, it would only be hardly possible to collect this sort of data. Hence, the manual definition of waterlines by human experts is the only possibility to derive ground truth references. However, the task of defining a waterline in an image is subject to individual perception to some extent. As a result there often is no unique solution but rather different experts will provide several different waterline estimates for the same image. Hence, there is a narrow range within which waterlines defined by different experts can be expected. Since our goal is to provide a method that mimics an expert when solving the regression problem to define a waterline, it is necessary to answer the question whether  the predicted waterline is in accordance with this narrow range of ambiguity. To put it more simple, the question should be answered whether a waterline prediction would be accepted by experts. We accounted for this and performed an evaluation as follows.

First, we assessed the quality of canoe segmentation as well as the inherently related classification of the particular disciplines (see Sec.~\ref{subsec:canoe_eval}). Second, we conducted a small study among several experts in the field of kinematic parameter analysis in canoe and kayak sprint (see Sec.~\ref{subsec:waterline_eval}). The study was the basis to define a ground truth reference as well as to quantify the uncertainty among experts. We subsequently assessed the accuracy of our predictions with respect to this gold standard.

\subsection{Canoe Segmentation and Classification}
\label{subsec:canoe_eval}
The segmentation of the canoe is important for the subsequent waterline prediction. We assessed the segmentation quality of the adjusted and trained Mask R-CNN using the validation set according to a standard evaluation metric. We used the intersection-over-union ($\mathit{IoU}$) defined as
\begin{align}
\mathit{IoU}=\frac{M^\prime \cap M}{M^\prime \cup M}
\end{align}
to measure the overlap between the predicted segmentation $M^\prime \in \{0,1\}^{m\times n}$ and the ground truth mask $M \in \{0,1\}^{m\times n}$ of the canoe. The former was selected from the predicted objects $\Omega$ as the one with the highest confidence value $p_v$.

We also assessed the classification performance of the algorithm. To that end, we evaluated whether the algorithm predicts the disciplines canoe sprint and kayak sprint correctly. We therefore determined the true and false positives/negatives on the validation set separately for each discipline and used them to calculate the corresponding F1 scores.

\subsection{Waterline Detection}
\label{subsec:waterline_eval}
%\subsubsection{Problem definition / Ambiguity problem and Evaluation Procedure}
The algorithm proposed in this work predicts the course of a waterline. Here, we present our evaluation procedure as well as the necessary preliminaries.

The parameterization of the waterline and an evaluation metric is presented in Sec.~\ref{subsec:waterline_eval_metric}. An evaluation study that was carried out to derive individual annotations (\ie, the determination of a waterline) for each image in the test set from several human experts is presented in Sec.~\ref{subsubsec:evalstudy}. Finally, the actual derivation of the ground truth reference data and the ambiguity between different expert annotations that can finally be used for the performance analysis of the algorithm are presented in Sec.~\ref{subsubsec:goldstandard}.

\subsubsection{Parameterization and Evaluation Metric}
\label{subsec:waterline_eval_metric}

\begin{figure}[t]
\begin{center}
\includegraphics[width=\linewidth]{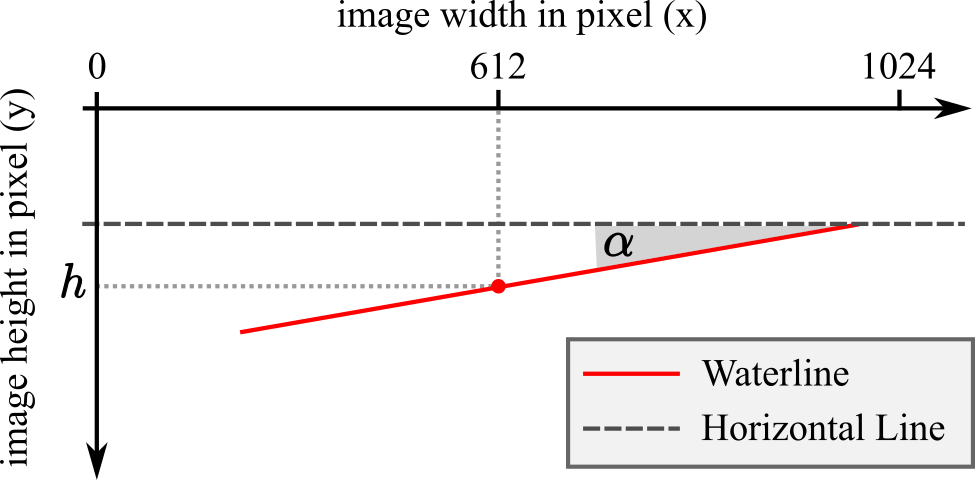}
\end{center}
\caption{Waterline parameterization: $h$ defines as the y-coordinate of the waterline at the center location in x-direction ($x=612$); $\alpha$ is the angle between waterline and horizontal line.}
\label{fig:waterline_parameters}
\end{figure}

The evaluation of a predicted waterline requires a suitable parameterization in order to perform comparisons with a reference.
Each waterline is a linear function and is clearly defined by its slope and a bias, \ie an interception with the coordinate axis at $x=0$. Without the loss of generality, we used a different parameterization which is shown in Fig.~\ref{fig:waterline_parameters}. It allows a better interpretation and comparison of the evaluation results. Firstly, it consists of a height parameter $h$ which is defined as the y-coordinate of the waterline at the center postion of the image (in x-direction), thereby effectively representing the location of the line. Secondly, an angle $\alpha$ which defines the rotation of the waterline with respect to the horizontal line is another parameter.

Note that this parameterization implies a rough similarity between the  waterlines in different images being evaluated. The consistence with respect to their position at and their extent along the x-direction  is particularly important to derive an estimate for the uncertainty between experts (see below). Due to the reasonably controlled conditions during video recording in our particular application scenario, this convention can be considered as true.

Given a particular waterline, the deviation between a given ground truth height $h_i$ and angle $\alpha_i$ and the predicted parameters $h^\prime_i$ and $\alpha^\prime_i$ in the $i$-th image defines as
\begin{align}
\epsilon^h_i = \left| h_i - h^\prime_i \right|
\label{eq:error_h}
\end{align}
\begin{align}
\epsilon^\alpha_i = \left| \alpha_i - \alpha^\prime_i \right|
\label{eq:error_alpha}
\end{align}
The actual definition of a suitable ground truth for the parameters $h_i$ and $\alpha_i$ for each image is subject of the evaluation study presented below.
%along with the estimated uncertainties defined by the boundaries $j$

\subsubsection{Evaluation study}
\label{subsubsec:evalstudy}
As mentioned before, a ground truth for the waterlines can only be defined on the basis of manual annotations. We conducted an evaluation study to derive multiple human annotations for each image in our test set for waterline evaluation. Therefore, we asked several experts from the field of kinematic parameter analysis in canoe sprint to determine a waterline. The implementation details were as follows.

The study was implemented on the basis of an interactive website. Given a test set (see below), we presented each image together with an initial guess of the waterline to each expert. The task of each expert was to carefully review the presented waterline and afterwards either modify its position and orientation by means of moving anchors at its ends or to accept the guessed line without any changes. Thereby, we controlled the initial guess as described below to prevent habituation to accept guessed lines without extensive review.

A total number of 130 images were selected from 66 videos to construct a test dataset $T$. 44 images were from canoe sprint and 86 from kayak sprint. We used these images to construct 4 groups within the test set. Group A: 90 images, the waterline as predicted by the algorithm without further modifications; Group B: 20 images drawn from group A, an additional offset of $-3$~pixels was added to the waterline; Group C: 10 images, a vertical shift of $+2$~pixels and a $-1.5\,^\circ$ rotation was added to the line as predicted by the algorithm; Group D: 10 images, similar processing as in group C, but with the rotation into the opposite direction, \ie $+1.5\,^\circ$. The groups B-D were used to enforce a misalignment of the presented waterline so that the participants are expected to perform modifications. The size of the distortions were selected by means of explorative tests to achieve variations that are visible but not trivial. The images were presented in a random order and without disclosing the group.

\subsubsection{Ground Truth and Ambiguity of Waterlines}
%Derivation of a Gold Standard
\label{subsubsec:goldstandard}

The dataset $T$ resulting from the evaluation study comprised multiple individual annotations for each image. We exploited this information to determine a ground truth waterline for each image as well as an estimate for the variation of the experts annotations as follows.

First, provided the individual parameters $h_{i,k}$ und $\alpha_{i,k}$, with $k\in N_i$ and  $N_i$ being the set of the individual experts, we calculated the mean value for each of the two parameters for each image $i, i\in T$, \ie
\begin{align}
\label{eq:h_i}
%h_i = \frac{1}{N^{\mathit{\asterisk}}_i}\cdot\sum_{k\in{N_i^*}}{h_{i,k}}
h_i = 1\left/ \left|N_i\right| \right. \cdot \sum_{k\in{N_i}}{h_{i,k}}
\end{align}
\begin{align}
\label{eq:alpha_i}
%h_i = \frac{1}{N^{\mathit{\asterisk}}_i}\cdot\sum_{k\in{N_i^*}}{h_{i,k}}
\alpha_i = 1\left/ \left|N_i\right| \right. \cdot \sum_{k\in{N_i}}{\alpha_{i,k}}
\end{align}
$\left|N_i\right|$ denotes the number of experts who annotated the $i$-th image. Next, the deviation of each individual parameter $h_{i,k}$ and $\alpha_{i,k}$ to the corresponding mean values $h_i$ and $\alpha_i$ were calculated by means of
\begin{align}
%h_i = \frac{1}{N^{\mathit{\asterisk}}_i}\cdot\sum_{k\in{N_i^*}}{h_{i,k}}
\epsilon^h_{i,k} =  h_{i,k} - h_i
\label{eq:delta_h}
\end{align}
\begin{align}
\epsilon^\alpha_{i,k} =  \alpha_{i,k} - \alpha_i
\label{eq:delta_alpha}
\end{align}
We employed these differences for further statistical analysis. We were particularly interested in the question whether there is statistical evidence that the individual annotations provided by different experts are similar and can therefore be used to calculate an average annotation for each image as introduced before in Eqs.~\ref{eq:h_i} and \ref{eq:alpha_i}. We applied a Kruskal-Wallis test as a non-parametric method to compare the distributions of the individual differences to the ground truth estimates, separately for the height and rotation parameter. The null hypothesis is that the medians of these  distributions are equal which would support the assumption that they originate from the same population. If the null hypothesis cannot be rejected in the light of the data, the parameters $h_i$ and $\alpha_i$ as introduced above seem a plausible approximation for the ground truth in each image. Hence, they can be used as the reference for further performance analysis.

Given this ground truth for each image, we were still interested in the overall variation of the individual annotations. Based on the individual deviations according to Eqs.~\ref{eq:h_i} and \ref{eq:alpha_i} for the entire dataset, we calculated the standard deviation for both waterline parameters, \ie
\begin{align}
\sigma_h = \sigma\left(\epsilon^h_{i,k}\right) \mathit{   } \forall \mathit{ } k \in N_i, i \in T \mathrm{, and}
\label{eq:sigma_h}
\end{align}
\begin{align}
\sigma_\alpha = \sigma\left(\epsilon^\alpha_{i,k}\right) \mathit{   } \forall \mathit{ } k \in N_i, i \in T\mathrm{,}
\label{eq:sigma_alpha}
\end{align}
These estimates serve as a general measure for the uncertainty among all experts. Using these measures, we finally defined an acceptance range, that means an interval in the vicinity of the ground truth reference parameters $h_{i}$ and $\alpha_{i}$, within which a predicted waterline would be considered as a valid estimate. This interval was constructed as the $u$-fold $\sigma_h$ and $\sigma_\alpha$ vicinity,
\begin{align}
\Delta h = \pm u \cdot \sigma_h
\label{eq:delta_h}
\end{align}
\begin{align}
\Delta \alpha = \pm u \cdot \sigma_\alpha
\label{eq:delta_alpha}
\end{align}
The parameter $u$ was determined such that $95\,\%$ of all individual annotations are contained within this range.

%the joint z-normalized distributions

%------------------------------------------------------------------------
\section{Results}
\begin{figure*}[t]
\begin{center}
\includegraphics[width=\linewidth]{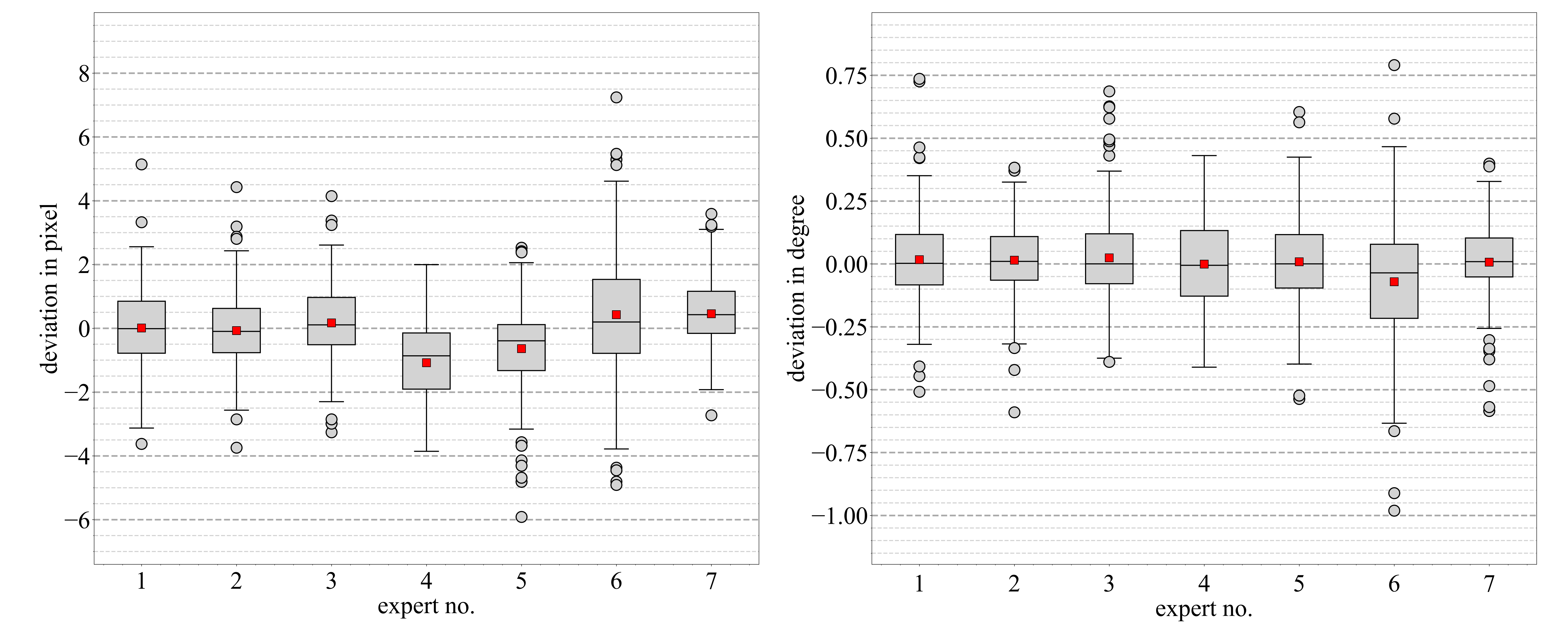}
\end{center}
   \caption{Distributions of the individual deviations from the estimated ground truth references shown for the height (left) and the rotation parameter (right). Upper / lower boundaries of boxes correspond to upper / lower quartiles. Whiskers denote the $1.5$-fold of the inter quartile range; circles denote outliers; vertical bars / red squares inside the boxes denote median / mean values.}
\label{fig:eval_experts}
\end{figure*}

\subsection{Canoe Segmentation and Classification}
The segmentation quality on the test dataset measured in terms of the $\mathit{IoU}$ was $0.82$ on average with a standard deviation of $0.04$ and minimal and maximal values of $0.72$ and $0.88$, respectively. This corresponds to a moderate and, more importantly, very consistent segmentation quality. The results clearly indicate a strong overlap between the true and the predicted masks of the canoe, which is particularly important for the subsequent waterline estimation.

The result of the classification performance analysis revealed a F1 score of $1.0$ for both disciplines. It means that the Mask R-CNN was perfectly able to distinguish between canoe and kayak sprint. The classification performance itself is only less important for the subsequent estimation of the waterline. However, it might be utilized in an automated processing pipeline to discriminate different paths in the analysis procedure which depends on the particular discipline.

\subsection{Ground Truth Waterline Parameters}
%(- gruppe 4 wurde weggelassen! -> bedeuten 6*20 Bilder weniger: -> 948 -> 828 Datensätze)
A total number of 7 experts participated in our study, 6 of which processed all 130 images and one processed only 48. This led to a total of 828 annotated waterlines in the test dataset. The distributions of the individual deviations from the ground truth as determined according to Eqs.~\ref{eq:delta_h} and \ref{eq:delta_alpha} are shown in Fig.~\ref{fig:eval_experts}. The visual comparison of the distributions reveals a general consent between experts. The rotation parameters are in stronger accordance to each other compared to the height parameters. This is also reflected in the upper and lower quartiles of these distributions, which are located within a range of only $\pm 0.22\,^\circ$ for the rotation and $\pm 1.96\,\mathit{px}$ for the height parameter.
Note the minor different appearance of the results for participants 4, 5 and 7 for the height parameter and for subject 6 for the rotation parameter compared to the other participants. These distributions are shifted slightly downwards for the height parameter for the participants 4 and 5 and upwards for participant 7. Regarding the rotation, the distributions standard deviation for participant 6 is considerably larger compared to all others.

The Kruskal-Wallis test was performed separately for each parameter given the null hypothesis that the medians of the distributions are similar using a significance level of $p=0.05$. We obtained the p-values $p=0.65$ for the rotation and $p=0.99$ for the height parameter. This indicates that the null hypothesis cannot be rejected in the light of the data. From this we concluded to estimate the ground truth reference for each waterline from the individual annotations provided by experts according to Eqs.~\ref{eq:h_i} and \ref{eq:alpha_i} separately for each image.

Next, we determined the standard deviation of the expert annotations according to Eqs.~\ref{eq:sigma_h} and \ref{eq:sigma_alpha} to $\sigma_h = 1.48\,\mathit{px}$ for the height parameter and $\sigma_\alpha = 0.20\,^\circ$ for the rotation parameter. Finally, we used Eqs.~\ref{eq:delta_h} and \ref{eq:delta_alpha} to estimate $u=2.5$ for the $u$-fold $\sigma_h$ and $\sigma_\alpha$ vicinity around the ground truth parameters $h_i$ and $\alpha_i$, provided the assumption that 95\,\% of the individual expert annotations should be contained in this vicinity for both height and angle parameter. The resulting intervals are $\Delta h = \pm 3.70\,\mathit{px}$ and $\Delta \alpha = \pm 0.50\,^\circ$. The tolerance interval for the height parameter corresponds to less than  $\pm 0.7\,\%$ of the spatial resolution (height dimension) of the images in the dataset.

%\subsection{Detection Accuracy of the Proposed Approach}
\subsection{Accuracy of Predicted Waterlines}
\begin{figure}[b]
\begin{center}
\includegraphics[width=\linewidth]{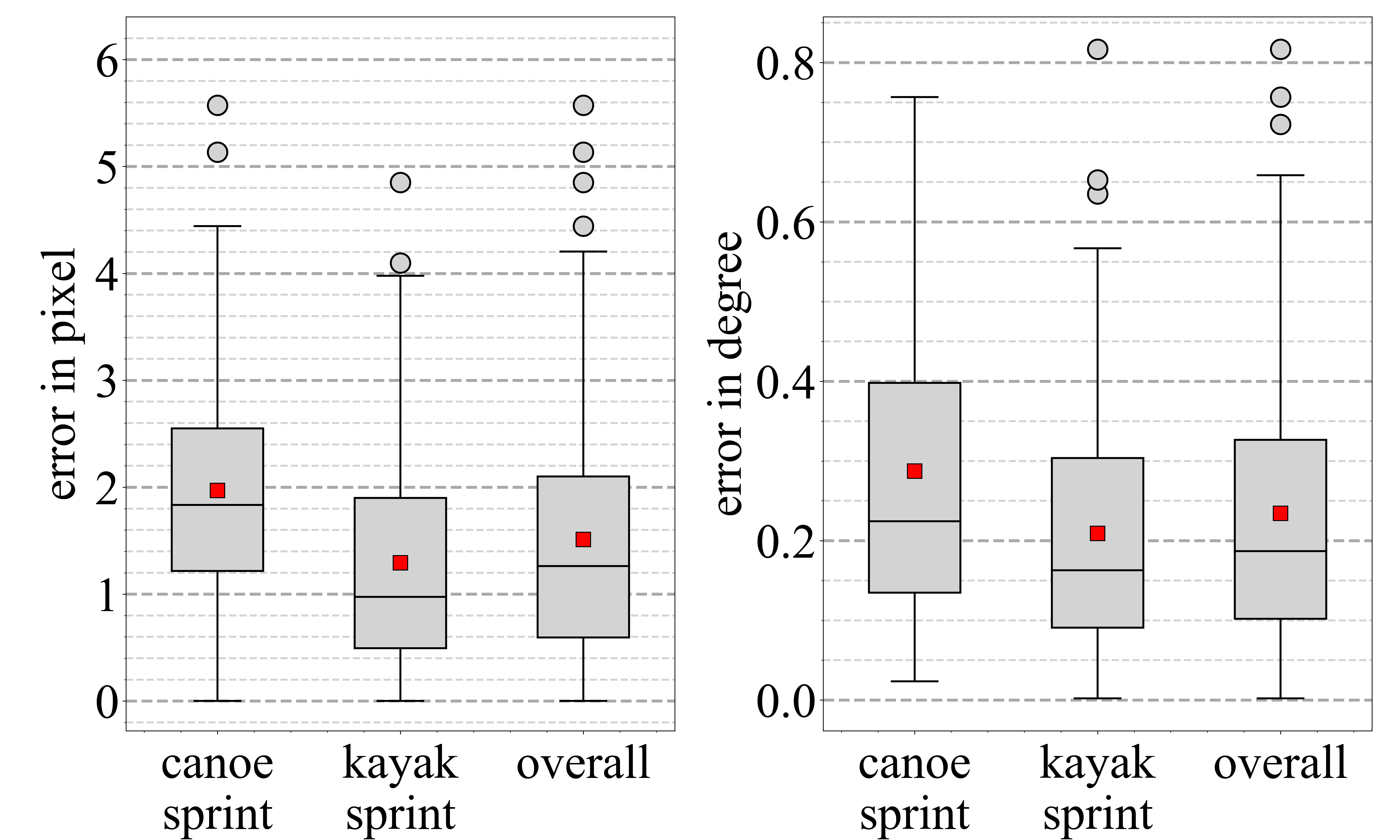}
\end{center}
   \caption{Error of predicted waterlines for height (left) and rotation parameter (right). Upper / lower boundaries of boxes correspond to upper / lower quartiles. Whiskers denote the $1.5$-fold of the inter quartile range; circles denote outliers; vertical bars / red squares inside the boxes denote median / mean values. See text for details.}
\label{fig:eval_predictions}
\end{figure}

We assessed the accuracy of the predicted waterlines by calculating the absolute differences to the ground truth parameters by means of Eqs.~\ref{eq:error_h} and \ref{eq:error_alpha}. The results for each discipline and the combined results are shown in Fig.~\ref{fig:eval_predictions}. It is obvious that the results obtained for canoe sprint appear to be slightly worse than for kayak sprint. Besides that it is shown that 50\,\% of the absolute differences for both height and rotation are less or equal than $1.26\,\mathit{px}$ and $0.19\,^\circ$, respectively, considering the distributions from the combined results. The largest values given the error metric are $5.57\,\mathit{px}$ for the height and $0.82\,^\circ$ for the rotation parameter.

Finally, we applied the $u$-fold $\sigma_h$ and $\sigma_\alpha$ vicinity as determined in the previous section to assess whether a waterline can be considered as a valid expert estimate. It turns out that a total number of 85\,\% of all predicted waterlines are in accordance with this interval. If the parameters are considered separately, even 95\,\% of the results for the rotation and 89\,\% for the height parameter fall into these intervals.

%------------------------------------------------------------------------
\section{Discussion and Conclusions}
We introduced an approach for the automatic detection of the waterline in canoe sprint video analysis. Our general goal was to provide an estimate for the course of waterlines in an image that could also have been defined by a human expert or, in other words, that experts would accept this prediction as a valid estimate. We achieved this for 85\,\% of the images in the validation dataset. Our solution for this particular regression problem comprises the segmentation of canoes based on an adjusted Mask R-CNN approach and a subsequent multi-stage procedure to estimate a waterline. We demonstrated its performance on a real dataset for which the ground truth references were derived on the basis of human expert annotations. Our solution provided robust and accurate results while still leaving space for further improvements and optimizations. They might not only refer to the segmentation part and the waterline estimation procedure but also to the derivation of the ground truth references.

Defining a suitable reference that can be used for further performance evaluations is a particular problem of this kind of regression tasks, \ie for which  a ground truth cannot be determined otherwise, \eg by means of sensors. In fact, the actual problem here is that the definition of a waterline is an ambiguous problem (caused by waves, splashes, a.s.o.), since it is prone to errors due to individual perception. As a consequence the results obtained from several experts for the same test image are subject to small variations. Hence, the ground truth can only be defined on the basis of an average result from several experts. Moreover, it is necessary to determine the amount of the variation between individual annotations. In fact, only the latter provides an actual meaning for the predicted waterline during evaluation, namely that a prediction is in accordance with the consent of experts. The quantification of these variations needs to be done carefully.

Here, we derived a ground truth and an estimate of the variation by means of a study among a small group of experts. This is certainly a limitation of our procedure, since it led to only a rather small number of annotations per image and, therefore, might have reduced the validity of the ground truth references. However, we assume that the domain knowledge of experts to define the waterlines implies that their individual deviation from the average annotation is small at all, and so the mean value of their annotations can be considered an appropriate estimate. The calculation of the overall ambiguity among experts is less affected since it was derived from the distribution of all deviations in the dataset rather than from individual images. Nonetheless, our gold standard definition is only valid with respect to the experimental conditions of the video recordings, \eg the perspective of the camera, the distance to the canoe and the spatial resolution of the images.

As mentioned before, 85\,\% of the predicted waterlines were in accordance with the gold standard we derived from the evaluation study and so 15\,\% were not. Importantly, our results show that the magnitudes of these outliers are still moderate. It is obvious that these proportions depend upon the definition of the $u$-fold $\sigma_h$ and $\sigma_\alpha$ vicinities. Here, it was selected such that 95\,\% of individual annotations were part of this interval. Less strict assumptions would of course improve the success rate. Further data is needed to derive a more sophisticated selection for this value. Moreover, outlier samples are worth to be analyzed separately and in more detail in order to identify potential systematic errors and to achieve further optimizations.

We carried out a brief error analysis and found a clear pattern for images that were not in accordance with experts annotations. These waterlines were slightly shifted upwards in the frontal part of the canoes compared to the ground truth. There is reason to believe that this is caused by larger waves in the frontal area resulting from, \eg, the cyclic movement of the canoe in upward and downward direction which is inherent to these disciplines. As a result, the waves occlude visible parts of the canoes front which shifts the segmentations, their outlines and finally also the waterlines. A possible solution is to not only restrict the linear regression to more central areas of the canoe segmentation but also to improve the canoe segmentation itself.

The good quality of the segmentation performance of the adjusted Mask R-CNN is effectively reflected by a high average $\mathit{IoU}$ value. The very small standard deviation underpins its general robustness, although the amount of data available for training and validation was fairly small. Increasing the amount of data might improve the performance significantly. Further limitations in the current dataset are unbalanced proportions of the samples with respect to the movement direction of canoes and to the actual discipline, \ie canoe or kayak sprint. Moreover, the dataset does not contain any negativ samples, that means images without any canoe. However, this is only of minor relevance if it can be ensured that the algorithm is applied to application specific data.

The developed method provides an important component for future developments towards an automated derivation and analysis of kinematic parameters from video and image recordings in canoe sprint and kayak sprint. A straightforward extension is the application and optimization of algorithms for human pose detection, \eg OpenPose~\cite{cao2016openpose}, which can provide coordinates of key-points on limb and face of athletes. Assessing such key-point positions with respect to the waterline can be used to derive kinematic parameters comparable to those used in todays human analysis~\cite{Fuchs2018}. Another extention is the detection of the paddle, which is important for several reasons. First, it provides another reference for key-point positions and the subsequent kinematic parameter analysis.  Second, it provides information on the relative time in a paddle cycle if evaluated in comparision to the waterline. The Mask R-CNN approach might be applied to the task of paddle segmentation as well.

Finally, the combination of these approaches paves the way for new applications in which not only several images but rather an entire video sequence can be analyzed. This provides new opportunities as, \eg, the utilization of temporal filters on the extracted parameters to achieve more robust predictions in single images but also to exploit the dynamics of kinematic parameters for the biomechanical analysis.

%These approaches, however, für die ableitung weiterer parameter ist es dann notwendig, auch z.b. die perspektive zu berücksichtigen /  weniger vorannahmen treffen zu müssen That means treffen bestimmte vorannahmen über die beschaffenhet der daten, die aber für solche spezifischen anwendungsfälle sinnvoll sind.

%(2) the method that is only applicable by trained experts an be used by less scientific experienced trainers / (may be also with less time), i.e. the technique can be applied in youth .. as early as possible (3) Basis to derive new temporal technical parameters and even the analysis of complete training rides.

{\small
\bibliographystyle{ieee_fullname}
\bibliography{egbib}
}

\end{document}